\pdfoutput=1

\documentclass[11pt]{article}

\usepackage{EACL2023}

\usepackage{times}
\usepackage{latexsym}

\usepackage[T1]{fontenc}

\usepackage[utf8]{inputenc}

\usepackage{microtype}

\usepackage{inconsolata}

\usepackage{graphicx}
\usepackage{amsmath,amssymb,amsfonts}
\usepackage{algorithm}
\usepackage{algorithmicx}
\usepackage{algpseudocode}
\usepackage{textcomp}
\usepackage{xcolor}
\usepackage{amsthm}
\usepackage{booktabs}
\usepackage{multirow}
\usepackage{graphicx}
\usepackage{svg}

\usepackage{array}
\usepackage{booktabs}

\title{From Graphs to Hypergraphs: Enhancing Aspect-Based Sentiment Analysis via Multi-Level Relational Modeling}

\author{Omkar Mahesh Kashyap$^{1}$\thanks{\quad Equal Contribution}~, Padegal Amit$^{1}$\footnotemark[1], Madhav Kashyap$^{2}$\\
\textbf{Ashwini M Joshi$^{1}$, Shylaja SS$^{1}$}\\
$^{1}$PES University ~~~$^{2}$University of Washington\\
 \texttt{\{omkar.m.kashyap, padegal.amit\}@gmail.com}, \texttt{madhavmk@uw.edu},\\ 
 \texttt{\{ashwinimjoshi, shylaja.sharath\}@pes.edu}
\\
}

\begin{document}
\maketitle
\begin{abstract}
Aspect-Based Sentiment Analysis (ABSA) predicts sentiment polarity for specific aspect terms, a task made difficult by conflicting sentiments across aspects and the sparse context of short texts. Prior graph-based approaches model only pairwise dependencies, forcing them to construct multiple graphs for different relational views. These introduce redundancy, parameter overhead, and error propagation during fusion, limiting robustness in short-text, low-resource settings. We present HyperABSA, a dynamic hypergraph framework that induces aspect-opinion structures through sample-specific hierarchical clustering. To construct these hyperedges, we introduce a novel acceleration-fallback cutoff for hierarchical clustering, which adaptively determines the level of granularity. Experiments on three benchmarks (Lap14, Rest14, MAMS) show consistent improvements over strong graph baselines, with substantial gains when paired with RoBERTa backbones. These results position dynamic hypergraph construction as an efficient, powerful alternative for ABSA, with potential extensions to other short-text NLP tasks.
\end{abstract}

\section{Introduction}

Aspect-Based Sentiment Analysis (ABSA) is a family of tasks that predict sentiment polarity with respect to specific aspects in text \cite{zhang2022survey}. In this work, we focus on the most widely studied subtask of ABSA, Aspect Term Sentiment Analysis (ATSA), where given a sentence and one or more marked aspect terms, the goal is to predict the sentiment polarity for each aspect.

\begin{figure}[htbp]
    \centering
    \includegraphics[width=\columnwidth]{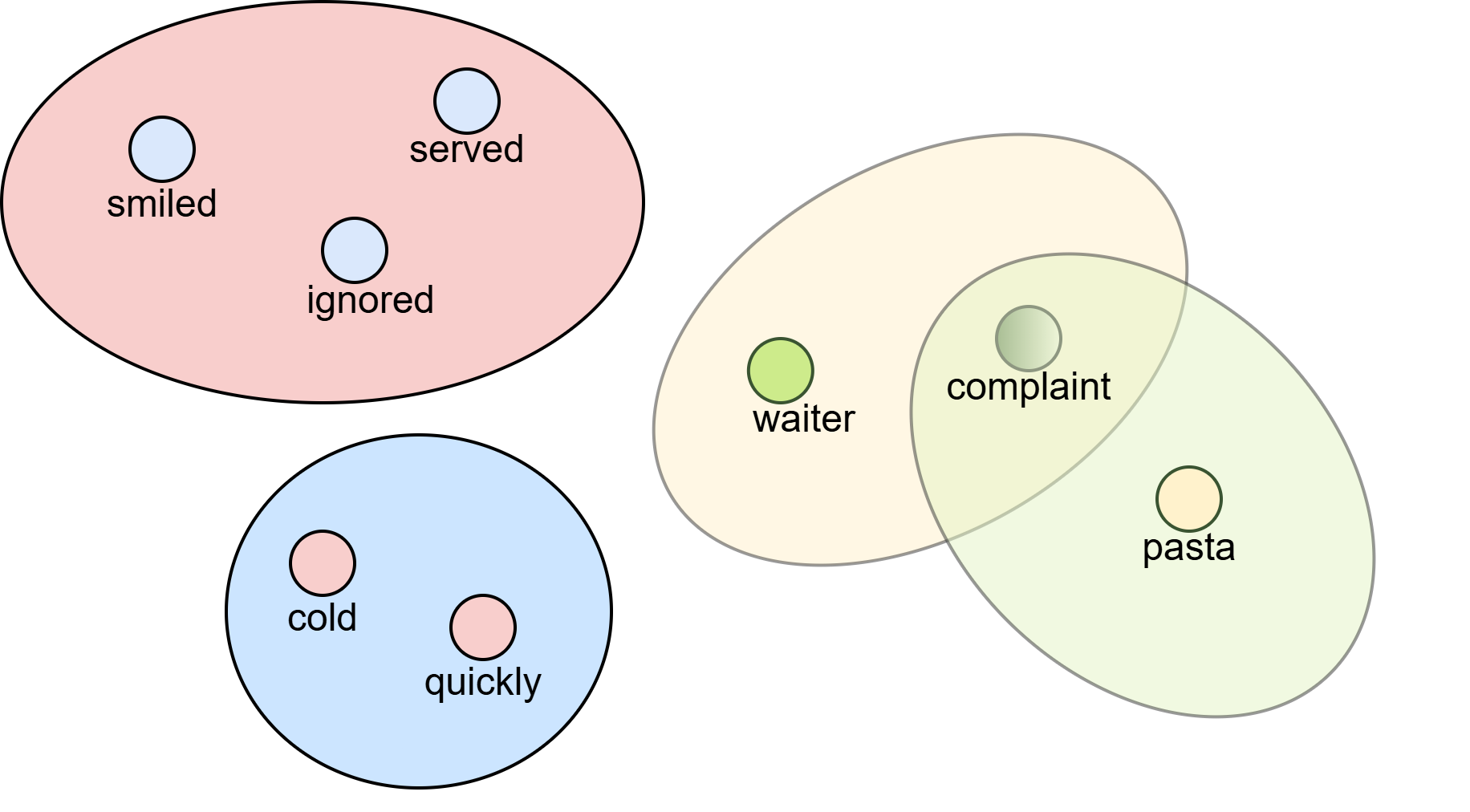} 
    \caption{A hypergraph of word interactions showing several semantic clusters based on aspect and sentiment polarity. This illustrates how words are grouped according to meaning and sentiment.}
    \label{fig:1}
\end{figure}

For instance, in the sentence \textit{``Service is good although a bit in your face , we were asked every five mins if food was ok, but better that than being ignored''}, the aspects ``service'' and ``food'' reflect positive and neutral sentiments, respectively. Such fine-grained opinion modeling is central to applications like product reviews, customer feedback, and social media monitoring.

Early work approached ATSA by incorporating dependency trees to capture syntactic relations between aspects and opinion words \cite{poria2014dependency, chen2022discrete}. More recently, graph neural networks (GCNs) have been widely adopted to propagate sentiment information along these structures \cite{kipf2016semi,  bai2020investigating, liang2020jointly, tian2021aspect, zhang2022ssegcn}. However, a fundamental limitation of these techniques lies in their inherent focus on pairwise relationships, because a single graph can only encode one type of relation at a time and cannot flexibly capture the multiple, overlapping relational views that matter in ABSA.

This gap has led to multi-graph architectures \cite{li2021dual, aziz2024unifying, zheng2024you}, which capture different facets of text, such as syntactic dependencies and semantic relationships, and then attempt to fuse information from these disparate graph sources.  While effective in fusing information, these methods introduce redundancy and parameter overhead,  require sophisticated fusion mechanisms and are vulnerable to error propagation from off-the-shelf syntactic parser outputs, particularly problematic in short texts where parser reliability is low. Such complexity makes them difficult to train robustly, especially in short-text, low-resource ABSA settings where data scarcity already limits generalization.

We address these challenges with HyperABSA, a dynamic hypergraph framework that unifies multiple graph views into a single adaptive structure. Unlike standard graphs that capture only pairwise relations, hypergraphs group tokens into hyperedges, directly modeling aspect–opinion structures and eliminating redundant fusion steps \cite{zhang2022hegel} (see Figure~\ref{fig:1}). Our key contribution is an adaptive hypergraph construction method tailored for ABSA. Each sentence undergoes hierarchical agglomerative clustering (HAC) over token embeddings, producing a dendrogram that is cut using an acceleration-fallback criterion. This criterion detects “elbows” in inter-cluster dissimilarity, signaling structural boundaries. When this signal is weak, a variance-sensitive threshold is used instead, ensuring robustness across sentence lengths. 

While second-order change detection is established in clustering, to our knowledge, this is the first application for per-instance hypergraph induction in ABSA. This enables sentence-specific structural modeling without relying on external parsers or fusion mechanisms. Experiments on Lap14, Rest14, and MAMS show consistent gains over strong graph baselines, achieving state-of-the-art performance with RoBERTa backbones.

This paper makes the following contributions to the field of ABSA:
\begin{itemize}

 \item We introduce HyperABSA, a hypergraph-based framework that replaces multi-graph fusion with a single adaptive structure, grouping aspects and opinion modifiers into hyperedges.

 \item We propose a per-instance dynamic hypergraph construction method, using hierarchical clustering with an acceleration-fallback cutoff that adapts to sentence length and structure.
 
 \item Our method achieves state-of-the-art performance on Lap14 and Rest14 with RoBERTa, and competitive results on MAMS. Extensive ablation studies validate the robustness of our design.

\end{itemize}

\section{Related Work}

Over the years, ABSA has been widely explored using various methodologies.

\subsection{Graph Based Methods}

Graph-based models are widely used in ABSA to encode syntactic and semantic relationships among tokens. Early approaches employed graph convolutional networks (GCNs) over dependency trees to model syntactic structure \cite{chen2019graph}, with later extensions incorporating both syntactic and semantic dependencies to enrich graph representations \cite{wang2020relational, chen2022discrete, bao2023exploring}.

To capture finer-grained relational cues, subsequent work introduced relational GCNs and type-aware architectures that explicitly model aspect-specific and inter-aspect dependencies \cite{huang2019syntax, yuan2020graph, tian2021aspect}. While effective, these approaches depend heavily on external syntactic parsers, which can be brittle in short, noisy sentences. Attention mechanisms were later integrated into graph models, with multi-head attention fused with GCNs to jointly encode syntactic and semantic signals \cite{xu2021attention, pan2023aspect, cui2023affective, zhang2024graph}.

More recent advances have explored heterogeneous graph structures to represent diverse linguistic relations while preserving their functional roles during sentiment propagation \cite{zeng2023aspect, niu2022composition}. In parallel, multi-graph frameworks construct complementary graph views to simultaneously encode localized aspect cues and global contextual semantics, enhancing sentiment inference across varied discourse structures \cite{li2021dual, aziz2024unifying, zheng2024you}.

\begin{figure*}[!t]
\centering
\includegraphics[]{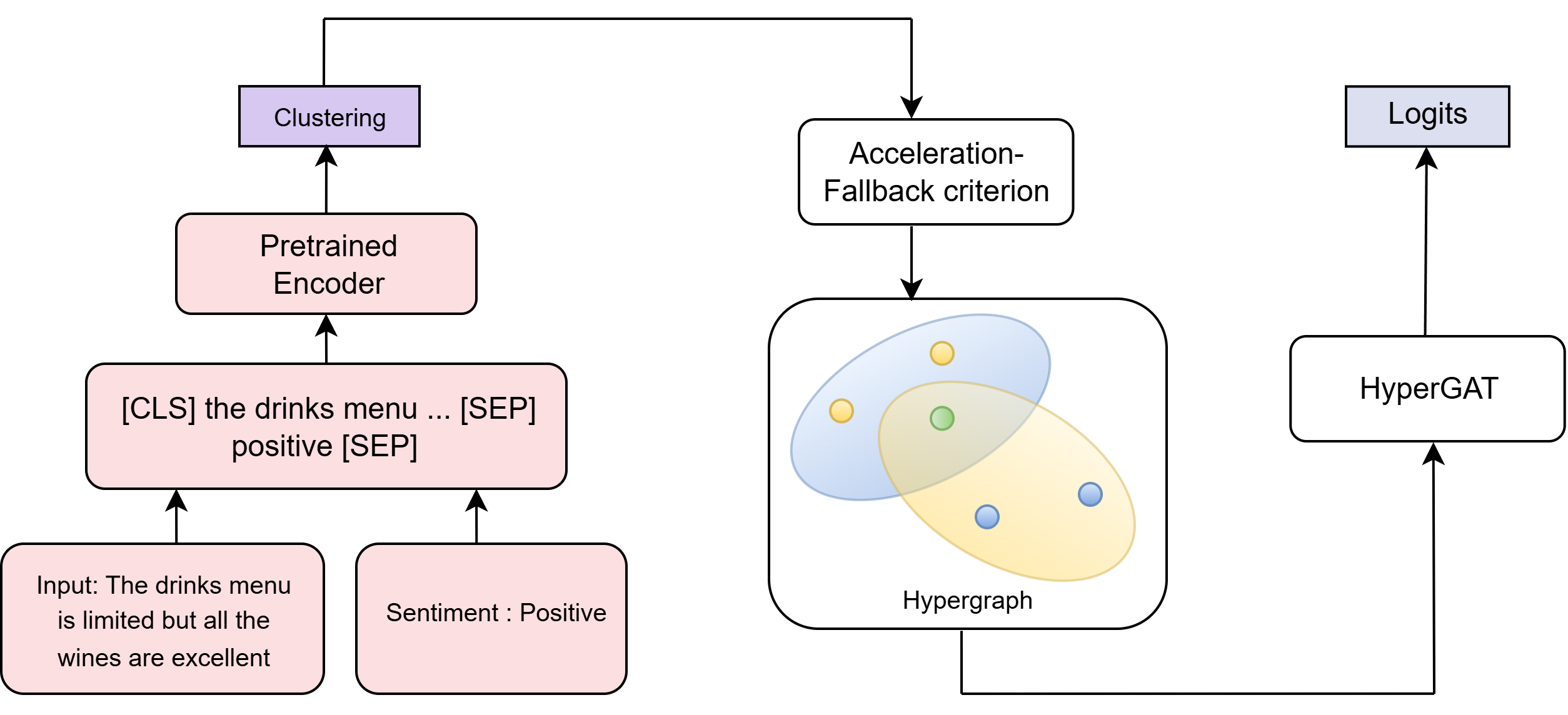}
\caption{Architecture of HyperABSA. \label{fig:2}}
\end{figure*}

\subsection{Hypergraph Construction Methods}

While hypergraph models have demonstrated strong potential across domains, their use in aspect-based sentiment analysis (ABSA) remains underexplored. Prior work largely advances hypergraph neural architectures \citep{feng2019hypergraph, zhi2024review}, with comparatively limited focus on text-driven hypergraph construction. Existing methods often connect tokens via nearest-neighbor relations in feature space \citep{yu2012adaptive, gao2022parallel, nguyen2020community, dai2023hypergraph}, which can introduce semantically irrelevant nodes. Topic-based strategies, such as LDA \citep{ding2020more, turnbull2024latent}, improve semantic grouping but depend on latent topic priors, while clustering-based approaches, like K-Means and spectral clustering enhance coherence at the cost of requiring fixed thresholds or cluster counts \citep{han1997clustering, chang2008unsupervised, saito2022hypergraph}.

Short-text ABSA presents additional challenges like sparse context, multiple aspects per sentence, and limited supervision, where global construction heuristics often fail. We address these issues with a per-instance adaptive cutoff that dynamically determines clustering depth, ensuring balanced granularity across sentences. Though inspired by classical elbow and change-point detection, our integration of acceleration-based criteria into per-instance hypergraph induction constitutes a novel methodological contribution with clear empirical gains.

\section{Methodology}

\subsection{Input Representation}

Given a sentence of $n$ tokens, we obtain contextual embeddings using a pretrained encoder: 
\[
\mathbf{X}^{(0)} = [\mathbf{x}_1, \ldots, \mathbf{x}_n] \in \mathbb{R}^{n \times d}
\]
where $\mathbf{x}_i \in \mathbb{R}^d$ is the contextual embedding of token $i$. These serve as initial node features for subsequent hypergraph construction.

\subsection{Hierarchical Clustering of Token Embeddings}

To induce semantic groupings, we apply HAC to the set \(
\{ \mathbf{x}_i \}_{i=1}^{n}
\), using a linkage function \( \ell(\cdot, \cdot) \). At each merge step \( t \in \{1, \ldots, n-1\} \), the two most similar clusters \( c_1, c_2 \) are merged, producing a new cluster of size \( s_t \) with dissimilarity \( \delta_t \).
\begin{align}
Z &= [z_1, \ldots, z_{n-1}] \in \mathbb{R}^{(n-1) \times 4} \nonumber \\
z_t &= [c_1, c_2, \delta_t, s_t]
\end{align}

The sequence \( \{ \delta_t \} \) encodes the evolution of inter-cluster dissimilarities.

\subsection{Adaptive Cutoff: Acceleration-Fallback Criterion}
A fixed global merge threshold may fail to account for variations in sentence length and syntactic complexity, potentially resulting in suboptimal clustering. To address this, we define an adaptive cutoff that selects an appropriate dendrogram height for each instance.

Let $\rho\in(0,1]$ control the fraction of merges considered. The number of merges in this window is:
\begin{equation}
r = \max\{1, \lfloor \rho \cdot (n-1)\rfloor\}.
\label{eq:adaptive_cutoff}
\end{equation}

Let \( \delta_{\text{recent}} = [\delta_{n - r}, \ldots, \delta_{n - 1}] \) denote the recent dissimilarities. 

If $|\boldsymbol{\delta}_{\mathrm{recent}}|>3$, we compute second-order finite differences (accelerations):
\begin{equation}
\kappa_j = \delta_{j+2} - 2\delta_{j+1} + \delta_j \quad \text{for } j = n - r, \dots, n - 3.
\end{equation}
to detect rapid changes in merge compactness and identify the elbow point:
\begin{equation}
j^\star = \arg\max_j \kappa_j
\end{equation}
However, if the recent window is too short (\( r \leq 3 \)), or the second-order signal is weak (\(
\max_j |\kappa_j| \leq \epsilon
\)), where \( \varepsilon\) is a small constant, situations common in sparse or noisy data where hierarchical structure is less distinct, we adopt a fallback cutoff: 
\begin{equation}
\delta_{\mathrm{fallback}} = \overline{\delta}_{\mathrm{recent}} + \lambda \cdot \sigma_{\mathrm{recent}}, \quad \lambda>0.
\label{eq:fallback_criterion}
\end{equation}

where $\overline{\delta}_{\text{recent}}$ and $\sigma_{\text{recent}}$ denote the mean and standard deviation of the recent dissimilarities, respectively and \(\lambda\) is a positive scaling parameter that controls how much the variability of recent dissimilarities influences the cutoff threshold.

The final dendrogram cutoff is then:

\begin{equation}
\delta_{\mathrm{elbow}} =
\begin{cases}
\min\{\delta_{j^*}, \delta_{\text{fallback}}\}, & 
\begin{aligned}
& \text{if } |\boldsymbol{\delta}_{\mathrm{recent}}|>3 \\
& \text{or } \max_j |\kappa_j| > \epsilon
\end{aligned}
\\[6pt]
\delta_{\text{fallback}}, & \text{otherwise}
\end{cases}
\label{eq:7}
\end{equation}

By cutting the dendrogram at the threshold at $\delta_{\text{elbow}}$, we obtain a partition of the data into $\lvert E \rvert$ clusters

We represent the clusters as hyperedges of a hypergraph $\mathcal{G} = (\mathcal{V}, \mathcal{E})$, where \( \mathcal{V} = \{1, \ldots, n\} \)
 corresponds to tokens, and \( E = \{ C_e \}_{e=1}^{|E|} \)
 denotes clusters obtained from the adaptive cutoff. The incidence matrix \( H \in \{0,1\}^{|V| \times |E|} \)
 is defined as:

\[
H_{v,e} =
\begin{cases}
1, & \text{if token } v \in C_e, \\
0, & \text{otherwise}.
\end{cases}
\]
The algorithm for our approach is presented in Appendix~\ref{sec:appendixA}.

\subsection{Hypergraph Neural Network}

Information is propagated in the hypergraph via a Hypergraph Attention Network (HyperGAT) \cite{ding2020more}.

For each attention head \( h \in \{1, \ldots, H\} \), the attention coefficients over nodes within each hyperedge are computed as:

\begin{equation}
A^{(h)} = \mathrm{softmax}\left(H^\top \cdot \sigma\big(X^{(0)} W_1^{(h)}\big)\right)
\end{equation}

where 
\(
W_1^{(h)} \in \mathbb{R}^{d \times d}
\)
is a learnable projection matrix for head \( h \), \( \sigma \) is an activation function and softmax is applied over the node dimension to normalize contributions within each hyperedge.

Each hyperedge \(e \in E
\) aggregates features from its incident nodes according to the learned attention weights:

\begin{equation}
E_{e,:}^{(h)} = \sum_{i=1}^n H_{i,e} \cdot A_{e,i}^{(h)} \cdot X^{(0)}_{i,:}
\end{equation}

This produces head-specific hyperedge representations \(E^{(h)} \in \mathbb{R}^{|E| \times d_h} \), where \(d_h\) denotes the dimensionality for the \(h\text{-th}\) attention head. The outputs from all \(H\) attention heads are concatenated along the feature dimension to form the final hyperedge features:
\begin{equation}
\begin{aligned}
E &= \text{Concat}\big(E^{(1)}, \ldots, E^{(H)}\big)
\end{aligned}
\label{eq:concat}
\end{equation}

To obtain a graph-level representation, we apply mean pooling over all hyperedges:
\[
\bar{E} = \frac{1}{|E|} \sum_{e=1}^{|E|} E_{e, :}
\]
The pooled embedding \(\bar{E}\) is then projected into the output label space through a linear classifier:
\[
\hat{y} = W_2 \bar{E} + b_2
\]

where \(W_2\) is a learnable linear projection matrix, \(b_2 \in \mathbb{R}^{C}\) is the bias term,  \( \mathcal{C} \) is the number of target classes, and \(d'\) is the dimensionality of the pooled embedding.

\subsection{Loss Function}

Given the pooled hyperedge representation 
\(\bar{E} \in \mathbb{R}^{d'}\), the model predicts a categorical distribution over the sentiment classes as:

\begin{equation}
P(y \mid \bar{E}) = \mathrm{softmax}(\hat{y}),
\label{eq:softmax}
\end{equation}
where $W_c \in \mathbb{R}^{|\mathcal{C}| \times d}$ and $b_c \in \mathbb{R}^{|\mathcal{C}|}$ are learnable parameters.

The overall loss function combines categorical cross-entropy loss with $\ell_2$ regularization:
\begin{equation}
\begin{split}
\mathcal{L} =\;& -\sum_{i=1}^{N} \sum_{c \in \mathcal{C}} \mathbf{1}(y_i = c) \log P(y_i = c \mid \bar{E}^{(i)}) \\
& + \beta \|\Theta\|_2^2,
\end{split}
\label{eq:loss}
\end{equation}

where $\Theta$ includes all trainable parameters, $\beta$ is the regularization weight and \(\mathbf{1}(\cdot)\) is the indicator function.

\begin{table*}[htbp]
    \centering
    \setlength{\tabcolsep}{2mm}  
    \begin{tabular}{lcc|cc|cc}
        \toprule
        \textbf{Model} & \multicolumn{2}{c|}{\textbf{Lap14}} & \multicolumn{2}{c|}{\textbf{Rest14}} & \multicolumn{2}{c}{\textbf{MAMS}} \\
        \cmidrule(lr){2-3} \cmidrule(lr){4-5} \cmidrule(lr){6-7}
                        & \textbf{Acc(\%)} & \textbf{F1(\%)} & \textbf{Acc(\%)} & \textbf{F1(\%)} & \textbf{Acc(\%)} & \textbf{F1(\%)} \\
        \midrule
        InterGCN\textsuperscript{\dag} \cite{liang2020jointly} & 78.06 & 73.83 & 85.45 & 77.64 
        & 82.49 & 81.95\\
        R-GAT\textsuperscript{\ddag} \cite{wang2020relational} & 78.21 & 74.07 & 84.64 & 77.14 & 83.16 & 82.42\\
        DGEDT\textsuperscript{\ddag} \cite{tang2020dependency} & 79.80 & 75.60   & 86.30 & 80.00  & - & -\\
        RGAT\textsuperscript{\ddag} \cite{bai2020investigating} & 80.31 & 76.38  & 85.77 & 79.81 & 82.96 & 82.12\\
        DMGLT \cite{fang2022dependencymerge} &78.82 & 75.56 & 86.25 & 79.04 &  - & - \\
        RMN \cite{zeng2022relation} &77.95 & 70.839 & 84.56 & 79.05 &   79.97 & 78.7\\
        CHGMAN\textsuperscript{\ddag} \cite{niu2022composition} & 78.04 & 74.46  & 85.98 & 79.31 & 83.23 & 82.66\\
        MWGCN \cite{yu2023novel} & 79.78 & 76.68 & 86.36 & 80.54 & - & - \\
        HGCN \cite{xu2023learn} & 79.59 & 76.24 & 86.45 & 80.60 & - & - \\
        UIKA-BERT\textsuperscript{\S}  \cite{liu2023unified} & 76.98 & 72.55 & 81.65 & 75.21 & - & - \\

        MTABSA-BERT\textsuperscript{\S}  \cite{zhao2023aspect} & 77.43 & 72.07 & 83.39 & 75.09 & - & - \\
        
        MLEGCN \cite{aziz2024unifying} & - & 69.74 & - & 79.94 & - & - \\
        SSIN \cite{wang2024multivariate} & 79.68 & 77.04 & 85.07 & 78.11 & - & - \\
        YORO\textsuperscript{\ddag}\cite{zheng2024you} & 77.45 & 73.21 &  83.69 &  76.22 & 84.21 & \textbf{83.78}  \\
        
        CaBiLSTM-BERT \cite{he2025cabilstm} & 77.91 & 73.04 & 83.75 & 75.87 & - & - \\

        \textbf{HyperABSA} &\textbf{83.44} & \textbf{80.60} & \textbf{87.76} & \textbf{82.76} & \textbf{84.56} & 83.74  \\

        \bottomrule
    \end{tabular}
    \caption{Performance of Accuracy and F1 score of HyperABSA with other models. \textsuperscript{\dag} denotes implementation from \cite{zheng2024you}, \textsuperscript{\ddag} denotes our implementation, and \textsuperscript{\S}  denotes implementation from \cite{he2025cabilstm} respectively}
    \label{tab:3}
\end{table*}

\begin{table*}[hbt!]
    \centering
    \setlength{\tabcolsep}{1mm}  
    \begin{tabular}{lc|cc|cc|cc}
        \toprule
        \textbf{Method Variant (with Formula)} & 
        \multicolumn{1}{c}{$\rho$} & 
        \multicolumn{2}{c}{\textbf{Rest14}} & 
        \multicolumn{2}{c}{\textbf{Lap14}} & 
        \multicolumn{2}{c}{\textbf{MAMS}} \\
        
        \cmidrule(lr){1-2} \cmidrule(lr){3-4} \cmidrule(lr){5-6} \cmidrule(lr){7-8}
        & & \textbf{ Acc. (\%)} & \textbf{ F1 (\%)} 
          & \textbf{ Acc. (\%)} & \textbf{ F1 (\%)} 
          & \textbf{ Acc. (\%)} & \textbf{ F1 (\%)} \\
        \midrule
        HyperABSA (Equation \ref{eq:7}) & Dynamic & \textbf{87.76} & \textbf{82.76} & \textbf{83.44} & \textbf{80.60} & \textbf{84.56} & \textbf{83.74} \\
        \midrule
        \(\delta_\text{elbow} = \delta_{fallback}\) & - & 84.07 & 76.89 & 79.06 & 75.84 & 84.00 & 83.51 \\
        \midrule
        \multirow{3}{*}{\(\delta_\text{elbow} = \delta_{n-m+k}\)} 
    & 0.2 & 80.59 & 74.61 & 79.68 & 76.30 & 83.48 & 82.82 \\   
    & 0.5 & 83.11 & 74.75 & 78.13 & 74.88 & 83.48 & 82.87 \\
    & 0.8 & 82.12 & 74.60 & 77.03 & 73.18 & 83.55 & 82.90 \\
\midrule
\multirow{3}{*}{\(\delta_\text{elbow} = \min(\delta_{n-m+k}, \delta_{fallback})\)} 
    & 0.2 & 84.78 & 77.35 & 79.22 & 77.14 & 84.22 & 83.46 \\ 
    & 0.5 & 80.95 & 72.06 & 78.75 & 75.95 & 84.07 & 83.24 \\
    & 0.8 & 84.98 & 78.24 & 79.53 & 76.03 & 83.70 & 83.09 \\

        \bottomrule
    \end{tabular}
    \caption{Ablation study on Rest14, Lap14, and MAMS showing the impact of acceleration formula and proportion ($p$) on HyperABSA’s performance. Formula types are indicated in parentheses within the method name.}
    \label{tab:4}
\end{table*}

\section{Experiments}

\subsection{Implementation Details}

For fair comparison, we follow the dataset pre-processing pipeline of RGAT \citep{wang2020relational}. Unlike RGAT, which depends on external syntactic features (POS tags, dependency relations, and head indices). Our model uses the full sentence, including all tokens such as stop words and aspect terms, as input to the encoder. These tokens, along with the encoder-specific tokens, are then used for hypergraph construction. Clustering is performed using Ward’s method with Euclidean distance on $\ell_2$-normalized token embeddings.

We evaluate HyperABSA on three standard ABSA benchmarks: MAMS \citep{jiang2019challenge}, SemEval-2014 Restaurants (Rest14), and Laptops (Lap14) \citep{pontiki-etal-2014-semeval}, following the data splits of \citet{bai2020investigating}. The text encoder is RoBERTa-base, chosen for its superior empirical performance. Dropout with rates in $[0.2, 0.3]$ is applied to both encoder and hypergraph layers, and $\ell_2$ regularization ($\beta = 2 \times 10^{-5}$) is used for weight decay. The clustering sparsity parameter $\rho$ and the variance-sensitivity coefficient $\lambda$ are treated as trainable parameters. Training is performed using the Adam optimizer \citep{kingma2014adam} with a learning rate of $10^{-2}$, a batch size of 16, and is conducted on a single NVIDIA RTX 4090 GPU.

\subsection{Baselines}
We benchmark HyperABSA against a comprehensive set of graph-based architectures leveraging syntactic dependencies and GCN variants, including InterGCN \cite{liang2020jointly}, R-GAT \cite{wang2020relational}, DGEDT \cite{tang2020dependency}, RGAT \cite{bai2020investigating}, DMGLT \cite{fang2022dependencymerge}, RMN \cite{zeng2022relation}, CHGMAN \cite{niu2022composition}, MWGCN \cite{yu2023novel}, HGCN \cite{xu2023learn}, UIKA-BERT \cite{liu2023unified}, MTABSA-BERT \cite{zhao2023aspect}, SSIN \cite{wang2024multivariate} and recent multi-graph models such as YORO \cite{zheng2024you}, MLEGCN \cite{aziz2024unifying} and CaBiLSTM-BERT \cite{he2025cabilstm}. To ensure fair comparison and reproducibility, we restrict evaluation to publicly available implementations.

\subsection{Results} 
As shown in Table \ref{tab:3}, HyperABSA consistently outperforms strong baselines across all three benchmark datasets. On the Lap14 dataset, it achieves a 7\% improvement in F1 score over CaBiLSTM-BERT, a recent 2025 baseline, and outperforms models such as YORO and RMN with an average gain of 5\% in accuracy. For the Rest14 dataset, HyperABSA obtains the highest accuracy and F1 score, surpassing YORO, MTABSA-BERT, R-GAT and RMN by an average margin of 4-5\%. On the more challenging MAMS dataset, HyperABSA outperforms InterGCN, DMGLT, and CHGMAN by an average of 2\% in both accuracy and F1, while remaining competitive with YORO in terms of F1 score. 
A comparison between HyperABSA and several LLM-based baselines on the Lap14 and Rest14 datasets is present in Appendix~\ref{sec:appendixD}

\section{Experimental Details}

\subsection{Comparison of Clustering Methods for Hypergraph Construction}
Table~\ref{tab:clustering-hyp} compares different clustering methods for hypergraph construction on Lap14 and Rest14 datasets. Density-based clustering (DBSCAN) \cite{ester1996density} performs poorly, likely due to the sparse and uneven distribution of aspect-opinion tokens, other methods such as GMM, KMeans, and random initialization achieve comparable accuracy and F1 scores, differing by only $\sim$0.1 in most cases. This consistency indicates that hypergraph-based ABSA is robust to minor variations in clustering quality. Notably, our variance-aware hierarchical clustering using Ward linkage \cite{murtagh2014ward} attains accuracy on par with or slightly above GMM and KMeans, while requiring significantly less computation. Although GMM and KMeans yield marginally higher scores in a few settings, their runtimes are substantially longer, as detailed in Appendix~\ref{sec:appendixB}.

\begin{table}[t]
\centering
\small
\setlength{\tabcolsep}{5pt}
\begin{tabular}{lcc|cc}
\toprule
\textbf{Model} & \multicolumn{2}{c|}{\textbf{Lap14}} & \multicolumn{2}{c}{\textbf{Rest14}} \\
              & Acc(\%) & F1(\%) & Acc(\%) & F1(\%) \\
\midrule
DBSCAN         & 47.30 & 27.86 & 52.87 & 47.92 \\
GMM            & 81.40 & 78.29 & \textbf{88.12} & 80.42 \\
KMeans         & 81.71 & \textbf{80.67} & 85.28 & 81.26 \\
Random         & 81.25 & 78.50 & 83.39 & 80.06 \\
No clustering  & 81.81 & 78.87 & 84.29 & 80.32 \\
\textbf{HyperABSA} & \textbf{83.44} & 80.60 & 87.77 & \textbf{82.76} \\
\bottomrule
\end{tabular}
\caption{Comparison of clustering methods evaluated by accuracy and F1 score on the Lap14 and Rest14 datasets}
\label{tab:clustering-hyp}
\end{table}

\subsection{Effect of Token-Level Preprocessing}

To evaluate the effect of token-level preprocessing on model performance, we perform ablations by selectively removing different categories of special tokens. Specifically, we remove stop words identified by the NLTK tokenizer \citep{bird2006nltk} before clustering, and test a variant that excludes tokenizer-specific special tokens such as \texttt{[CLS]} and \texttt{[SEP]} for BERT, and \texttt{<s>} and \texttt{</s>} for RoBERTa.

As shown in Table~\ref{tab:stop-words-ablation-study}, removing stop words substantially degrades performance across all datasets, highlighting their importance in maintaining syntactic and semantic coherence during hypergraph construction. Excluding sentence or tokenizer markers also reduces performance, though less severely. The best results are achieved when both stop words and special tokens are retained, suggesting that these contextual cues enable the clustering stage to capture clearer structural boundaries. Overall, token-level cues, often discarded in standard preprocessing, prove essential for stable hypergraph induction and improved sentiment prediction.

\begin{table}[htbp]
\centering
\resizebox{0.48\textwidth}{!}{
\begin{tabular}{lcc|cc|cc}
\toprule
 & \multicolumn{2}{c|}{\textbf{Lap14}} & \multicolumn{2}{c|}{\textbf{Rest14}} & \multicolumn{2}{c}{\textbf{MAMS}} \\
 & Acc(\%) & F1(\%) & Acc(\%) & F1(\%) & Acc(\%)& F1(\%)\\
\midrule
No special tokens      & 80.47 & 78.46 & 86.05 & 79.58 & 82.39 & 80.45 \\
No stop words          & 75.16 & 72.06 & 80.79 & 74.04 & 75.16 & 72.06 \\
Including both         & \textbf{83.44} & \textbf{80.60} & \textbf{87.76} & \textbf{82.76} & \textbf{84.56} & \textbf{83.74} \\
\bottomrule
\end{tabular}
}
\caption{Ablation study on the impact of token-level preprocessing strategies using the RoBERTa encoder}
\label{tab:stop-words-ablation-study}
\end{table}

\subsection{Embedding Analysis}

We evaluate the impact of different text encoders on HyperABSA to assess the robustness of our framework across backbone architectures. As shown in Table~\ref{tab:emb-ablation}, RoBERTa consistently outperforms BERT and XLNet across all three datasets (Lap14, Rest14, and MAMS) in both accuracy and F1 score. The improvements are particularly pronounced on the Rest14 dataset, indicating that RoBERTa’s stronger contextual representations enhance the quality of token embeddings used for hypergraph construction. 

\begin{table}[htbp]
\centering
\resizebox{0.48\textwidth}{!}{
\begin{tabular}{lcc|cc|cc}
\toprule
\textbf{Model} & \multicolumn{2}{c|}{\textbf{Lap14}} & \multicolumn{2}{c|}{\textbf{Rest14}} & \multicolumn{2}{c}{\textbf{MAMS}} \\
              & Acc(\%) & F1(\%) & Acc(\%) & F1(\%) & Acc(\%) & F1(\%) \\
\midrule
RoBERTa       & \textbf{83.44} & \textbf{80.60} & \textbf{87.76} & \textbf{82.76} & \textbf{84.56} & \textbf{83.74} \\
BERT          & 79.69 & 77 & 86.67 & 80 & 84.56 & 83 \\
XLNet         & 76.88 & 74 & 79.69 & 77 & 79.69 & 77 \\
\bottomrule
\end{tabular}
}
\caption{Accuracy and F1 scores of different encoders on ABSA datasets.}
\label{tab:emb-ablation}
\end{table}

In contrast, XLNet underperforms due to its permutation-based pretraining, which appears less suited for short, aspect-focused sentences typical in ABSA. These results confirm that HyperABSA benefits from high-quality encoder representations but remains flexible and effective across different pretrained backbones.

\subsection{Linkage and Distance ablation}

We analyze the effect of different linkage criteria and distance metrics used in hierarchical clustering for hypergraph construction. As shown in Table~\ref{tab:full-clustering-results}, Ward linkage with Euclidean distance achieves the highest accuracy and F1 across both Lap14 and Rest14 datasets, outperforming Single, Complete, and Average linkage methods. Ward's variance-minimizing criterion produces more coherent and balanced clusters, yielding interpretable and sentiment-aligned hyperedges. Single and Complete linkage tend to produce fragmented or overly rigid clusters, while Average linkage overly smooths boundaries, obscuring fine-grained sentiment cues. Although other distance metrics such as cosine perform competitively under Single, Complete, or Average linkage, Euclidean distance remains consistently superior due to its alignment with $\ell_2$-normalized token embeddings. Overall, these results confirm that variance-aware hierarchical clustering with Euclidean geometry (Ward + Euclidean) yields the most stable and semantically meaningful hypergraph structures for ABSA.

\begin{table*}[htbp]
\centering
\resizebox{\textwidth}{!}{%
\begin{tabular}{l|cc|cc|cc|cc|cc|cc|cc|cc}
\toprule
& \multicolumn{4}{c|}{\textbf{Single}} 
& \multicolumn{4}{c|}{\textbf{Complete}} 
& \multicolumn{4}{c|}{\textbf{Average}} 
& \multicolumn{4}{c}{\textbf{Ward}} \\
\cmidrule(lr){2-5} \cmidrule(lr){6-9} \cmidrule(lr){10-13} \cmidrule(lr){14-17}
& \multicolumn{2}{c|}{\textbf{Lap14}} & \multicolumn{2}{c|}{\textbf{Rest14}} 
& \multicolumn{2}{c|}{\textbf{Lap14}} & \multicolumn{2}{c|}{\textbf{Rest14}} 
& \multicolumn{2}{c|}{\textbf{Lap14}} & \multicolumn{2}{c|}{\textbf{Rest14}} 
& \multicolumn{2}{c|}{\textbf{Lap14}} & \multicolumn{2}{c}{\textbf{Rest14}} \\
\cmidrule(lr){2-3} \cmidrule(lr){4-5} \cmidrule(lr){6-7} \cmidrule(lr){8-9}
\cmidrule(lr){10-11} \cmidrule(lr){12-13} \cmidrule(lr){14-15} \cmidrule(lr){16-17}
\textbf{Metric} 
& \textbf{Acc} & \textbf{F1} 
& \textbf{Acc} & \textbf{F1} 
& \textbf{Acc} & \textbf{F1} 
& \textbf{Acc} & \textbf{F1} 
& \textbf{Acc} & \textbf{F1} 
& \textbf{Acc} & \textbf{F1} 
& \textbf{Acc} & \textbf{F1} 
& \textbf{Acc} & \textbf{F1} \\
\midrule
\textbf{Euclidean} 
& 81.45 & 78.85 & 86.51 & 80.52 
& 82.56 & 79.96 & 86.68 & 81.65 
& 82.34 & 79.35 & 83.44 & 80.60 
& 83.44 & 80.60 & 87.76 & 82.76 \\
\textbf{Cosine} 
& 81.88 & 79.49 & 84.10 & 79.89 
& 82.44 & 78.85 & 83.85 & 78.95 
& 81.23 & 79.87 & 84.20 & 80.67 
& 82.67 & 79.94 & - & - \\
\bottomrule
\end{tabular}
}
\caption{Clustering results on Lap14 and Rest14 datasets showing Accuracy and F1 scores (\%) under different linkage methods and distance metrics.}
\label{tab:full-clustering-results}
\end{table*}

\subsection{Cluster Quality Analysis}
        We evaluate the effectiveness of our hypergraph construction method by comparing it against (i) a Random hypergraph, where nodes and hyperedges are formed without structural guidance, and (ii) a KNN-KMeans hybrid hypergraph that combines local and global relational signals via K-Nearest Neighbors and K-Means clustering. Cluster quality is assessed using the Silhouette Score \citep{rousseeuw1987silhouettes}, which captures intra-cluster cohesion and inter-cluster separation (higher is better), and the Davies-Bouldin Score \citep{davies1979cluster}, which quantifies average inter-cluster similarity (lower is better), across different training epochs.

As shown in Table~\ref{tab:cluster-quality}, HyperABSA consistently outperforms both baselines. Random hypergraphs, due to their stochastic construction, fail to produce meaningful clusters, often resulting in negative Silhouette Scores. The KNN-KMeans hybrid, while incorporating local and global structural cues, still falls short of HyperABSA in clustering quality. These results highlight that our method preserves both structural and semantic coherence across training epochs. Results regarding the complexity of the construction process in Sec: Appendix.

\begin{figure}[htbp]
    \centering
    \includegraphics[width=\columnwidth]{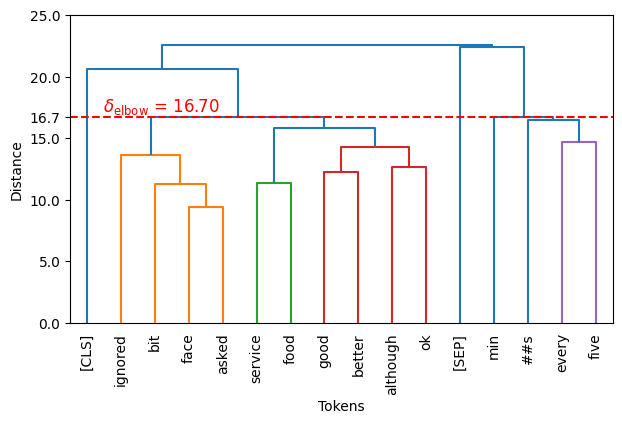} 
    \caption{Dendrogram visualization of token-level hierarchical clustering.}
    \label{fig:dendrogram-threshold}
\end{figure}

\begin{table}[t]
\centering
\resizebox{0.48\textwidth}{!}{
\begin{tabular}{lccc|ccc}
\toprule
\textbf{Model} 
& \multicolumn{3}{c|}{\textbf{Silhouette Score}} 
& \multicolumn{3}{c}{\textbf{Davies-Bouldin Score}} \\
\cmidrule(lr){2-4} \cmidrule(lr){5-7}
& \textbf{Min} & \textbf{Mean} & \textbf{Max} 
& \textbf{Min} & \textbf{Mean} & \textbf{Max} \\
\midrule
Random         & -0.24 & -0.23 & -0.22 & 1.51 & 1.59 & 1.64 \\
KMeans     &  0.31 &  0.33 &  0.40 & 1.05 & 1.17 & 1.32 \\
\textbf{HyperABSA} 
               & \textbf{0.36} & \textbf{0.42} & \textbf{0.62} 
               & \textbf{0.56} & \textbf{0.99} & \textbf{1.10} \\
\bottomrule
\end{tabular}
}
\caption{Comparison of cluster quality across different hypergraph construction methods.}
\label{tab:cluster-quality}
\end{table}

\subsection{Multi Granular Approach of Hypergraph}

To explore whether a dynamically constructed hypergraph can serve as a viable alternative to manually designed multi-graph architectures for multi-granular reasoning, we conduct a series of comparative experiments. We compare our dynamic hypergraph approach with several fixed-granularity baselines, including models with only fallback connections (coarse granularity), and acceleration paths with static thresholds ($\rho$ = 0.2, 0.5, 0.8). As seen in Table \ref{tab:4}, across datasets, these fixed strategies yield lower or inconsistent performance, indicating their inability to capture the optimal granularity across samples. In contrast, our model dynamically selects both the threshold and the graph construction strategy per instance, effectively adapting to sample-specific views. These findings support our broader claim, that automatically identifying an appropriate granularity per instance can offer a strong alternative to using multiple graphs for capturing the different granularities.

\subsection{Generalization Gap}

To rigorously evaluate the generalization capabilities of HyperABSA, we investigate the \textit{generalization gap}, defined as the difference between training and test accuracy and loss, across varying proportions of training data. Unlike prior works that rely on multiple graph constructions capturing distinct semantic views at the cost of increased complexity, our model leverages a single dynamic hypergraph, hypothesized to better balance representation richness and overfitting risk. We conduct controlled experiments on two widely-used benchmarks, Lap14 and Rest14, comparing HyperABSA against two strong baselines: YORO and RGAT. Each experiment is repeated over multiple random seeds, and average metrics are reported to ensure statistical reliability.

\begin{figure}[htbp]
    \centering
    \includegraphics[width=\columnwidth]{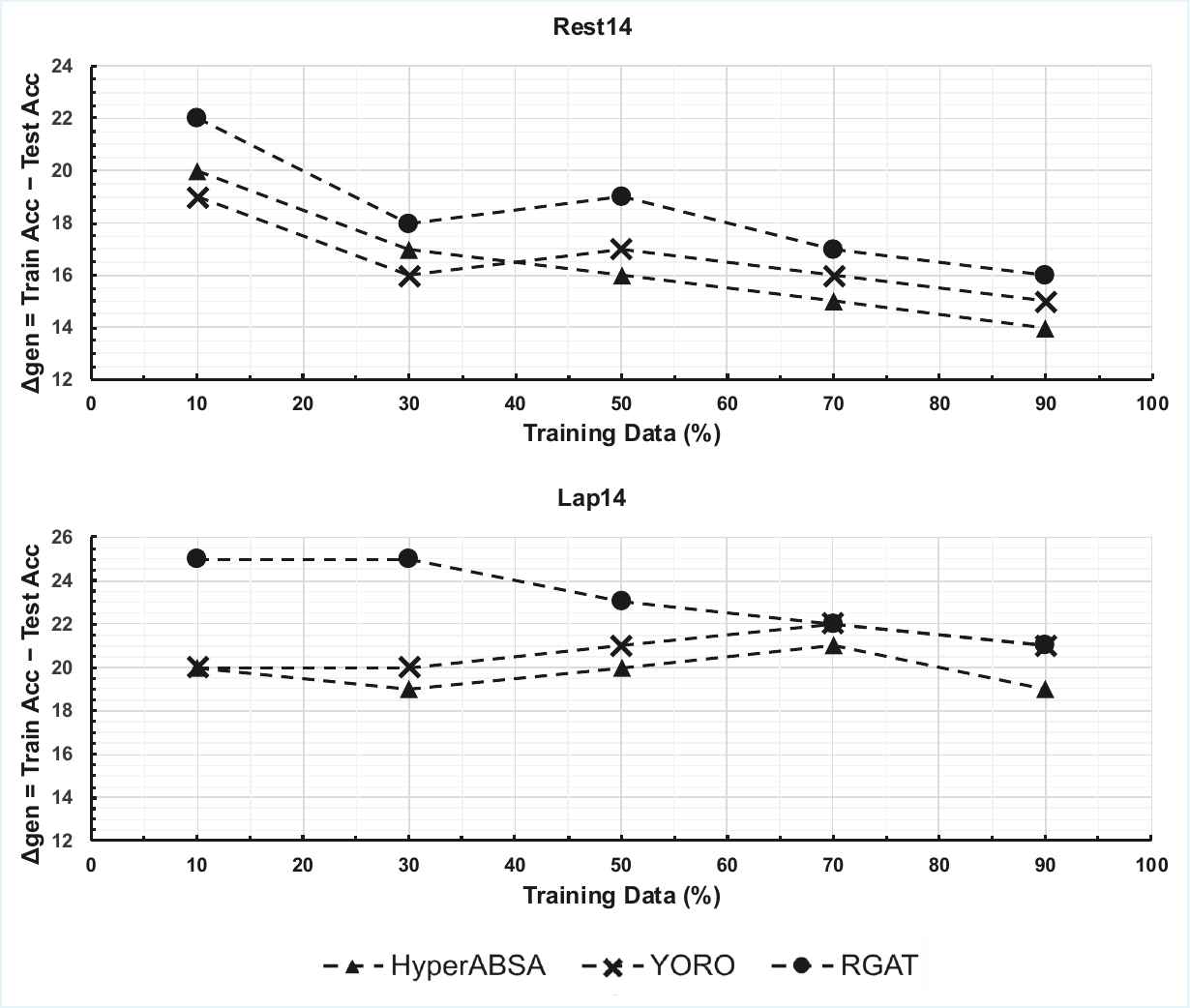}
    \caption{Evaluation of HyperABSA against multi-graph-based models on the Rest14 and MAMS datasets in terms of generalization gap.}
    \label{fig:7}
\end{figure}

Figure \ref{fig:7} illustrates that HyperABSA consistently demonstrates significantly smaller generalization gaps across almost all training data fractions. Notably, HyperABSA performs reliably even when trained on less than 50\% of the data, whereas both baselines require 50–70\% of the data to achieve comparable generalization. These trends suggest that the dynamic structure of HyperABSA enables it to capture fine-grained, context-sensitive token relationships without overfitting, even in low-resource settings.

\section{Conclusion}

In this paper, we introduce HyperABSA, a novel hypergraph construction methodology for ABSA that dynamically forms hyperedges via adaptive hierarchical clustering. Our approach addresses the challenge of overfitting in short-text scenarios by leveraging an efficient, acceleration-based thresholding mechanism, ensuring that hyperedges capture meaningful multi-node interactions while preventing excessive fragmentation or over-merging. Comprehensive evaluations on Lap14, Rest14, and MAMS datasets demonstrate that HyperABSA achieves state-of-the-art performance among graph-based approaches, highlighting its effectiveness in capturing nuanced multi-node interactions for fine-grained sentiment reasoning.

\section{Limitations}

Multi-graph models offer interpretable edge semantics grounded in syntactic or semantic roles, while hypergraphs, though rich in context, lack this clarity, posing challenges for interpretability and fine-grained error analysis. Our approach is computationally complex compared to conventional single-graph baselines, making it susceptible to overfitting, particularly on low-resource datasets such as Lap14, where aspect-opinion annotations are sparse and domain-specific vocabularies limit generalization. Although we introduced minor architectural adjustments to the base HGNN framework, it was not designed for ABSA. This mismatch added to the modeling complexity and may have hindered performance in ABSA-specific scenarios.



\clearpage
\onecolumn

\appendix

\section*{Appendix A \quad Algorithm of HyperABSA}
\label{sec:appendixA}

To formally describe our approach, Algorithm~\ref{alg:1} outlines our acceleration-based elbow method for hypergraph construction.

\begin{algorithm}[H]
\caption{Acceleration-Based Elbow Method for Hypergraph Construction}
\label{alg:1}
\begin{algorithmic}[1]
\Statex \textbf{Input:}  Linkage matrix $Z \in \mathbb{R}^{(n-1)\times 4}$ 
\Statex \textbf{Hyperparameters:} Proportion $\rho \in (0,1]$, fallback scale $\lambda>0$, tolerance $\epsilon>0$
\Statex \textbf{Output:} Cutoff threshold $\delta_{\mathrm{elbow}}$

\State $r \gets \max\!\big(1,\lfloor \rho\cdot (n-1)\rfloor\big)$
\State $\delta_{\mathrm{recent}} \gets [\delta_{n-r},\delta_{n-r+1},\ldots,\delta_{n-1}]$ extracted from $Z$

\If{$|\delta_{\mathrm{recent}}|>3$}
  \State \textbf{Compute second-order differences}
  \For{$j \gets n-r$ \textbf{to} $n-3$}
    \State $\kappa_j \gets \delta_{j+2}-2\delta_{j+1}+\delta_j$
  \EndFor
  \State $j^\star \gets \arg\max_j \kappa_j$
  \State $\delta_{\mathrm{elbow}}^{(1)} \gets \delta_{j^\star}$
\EndIf

\State \textbf{Compute fallback threshold}
\State $\bar{\delta}_{\mathrm{recent}} \gets \mathrm{mean}(\delta_{\mathrm{recent}})$
\State $\sigma_{\mathrm{recent}} \gets \mathrm{std}(\delta_{\mathrm{recent}})$
\State $\delta_{\mathrm{fallback}} \gets \bar{\delta}_{\mathrm{recent}} + \lambda\cdot \sigma_{\mathrm{recent}}$

\If{$|\delta_{\mathrm{recent}}|>3$ \textbf{ or } $\max_j |\kappa_j|>\epsilon$}
  \State $\delta_{\mathrm{elbow}} \gets \min\!\big(\delta_{\mathrm{elbow}}^{(1)},\,\delta_{\mathrm{fallback}}\big)$
\Else
  \State $\delta_{\mathrm{elbow}} \gets \delta_{\mathrm{fallback}}$
\EndIf

\State \Return $\delta_{\mathrm{elbow}}$
\end{algorithmic}
\end{algorithm}





\section*{Appendix B \quad Efficiency Analysis of HyperABSA}
\label{sec:appendixB}

The primary computational cost in HyperABSA arises from the Hierarchical Agglomerative Clustering (HAC) step, which constructs a dendrogram by comparing every pair of tokens in a sentence. If each sample contains \( n \) tokens, then HAC requires computing \( n^2 \) pairwise distances and merging clusters iteratively.  
Therefore, in the worst case, its time and memory complexity is \( \mathcal{O}(n^2) \).
However, since ABSA sentences are typically short ($n \leq 50$), this cost is very small in practice.

The adaptive cutoff mechanism adds only a minor overhead. It performs a single linear pass over the last $m$ merges in the dendrogram (as defined in Equation \eqref{eq:adaptive_cutoff}) to compute simple statistics such as the mean and standard deviation for detecting structural transitions. This operation costs at most $\mathcal{O}(m) \leq \mathcal{O}(n)$, which is insignificant relative to the clustering step.

For comparison, elbow-based K-Means methods are much more expensive. They need to run the clustering algorithm multiple times for different number of clusters \( K \in \{1, \dots, K_{\max}\}\). Each K-Means run has a complexity of $\mathcal{O}(ndT)$, where $n$ is the number of tokens, $d$ is the embedding dimension and $T$ denotes the number of iterations per run. Thus, the total cost becomes $\mathcal{O}(K_{\max} ndT)$, which grows quickly and makes K-Means unsuitable for per-instance (adaptive) clustering.

In summary, HyperABSA achieves per-sample adaptivity with minimal overhead.
It keeps the efficiency of HAC while avoiding repeated clustering runs, making it well-suited for real-time or token-level ABSA tasks.

\begin{table}[ht]
\centering
\begin{tabular}{lcc}
\toprule
\textbf{Method} & \textbf{Lap14 (min:secs)} & \textbf{Rest14 (min:secs)} \\
\midrule
No HG   & 9:37  & 14:39 \\
Random  & 10:30 & 20:20 \\
GMM     & 27:43 & 44:15 \\
KMeans  & 55:00 & 144:20 \\
HyperABSA   & 11:31 & 17:36 \\
\bottomrule
\end{tabular}
\caption{Comparison of times in minutes:seconds to run for different methods on the Lap14 and Rest14 datasets}
\label{tab:time-comparison}
\end{table}

Empirically (see Table \ref{tab:time-comparison}), our adaptive hypergraph construction is highly efficient compared to other clustering approaches. While it takes slightly longer than running with No Hypergraph ("No HG") and a randomly constructed Hypergraph (“Random”), due to the adaptive cutoff computation, the extra time is small and only about 1-2 minutes overall. In contrast, Gaussian Mixture Models (GMM) and K-Means take significantly longer because they repeatedly compute distances and refine clusters, resulting in much higher runtimes. They exceed ours by over 3x and 8x respectively. This demonstrates that HyperABSA balances adaptivity and efficiency while introducing meaningful structural modeling at negligible extra cost, while remaining far more scalable than conventional clustering approaches. All experiments were conducted on a single NVIDIA RTX 4090 GPU.

\section*{Appendix C \quad Interpretability of Hyperedges}
\label{sec:appendixC}

\begin{figure}[htbp]
    \centering
    \includegraphics[width=0.5\columnwidth]{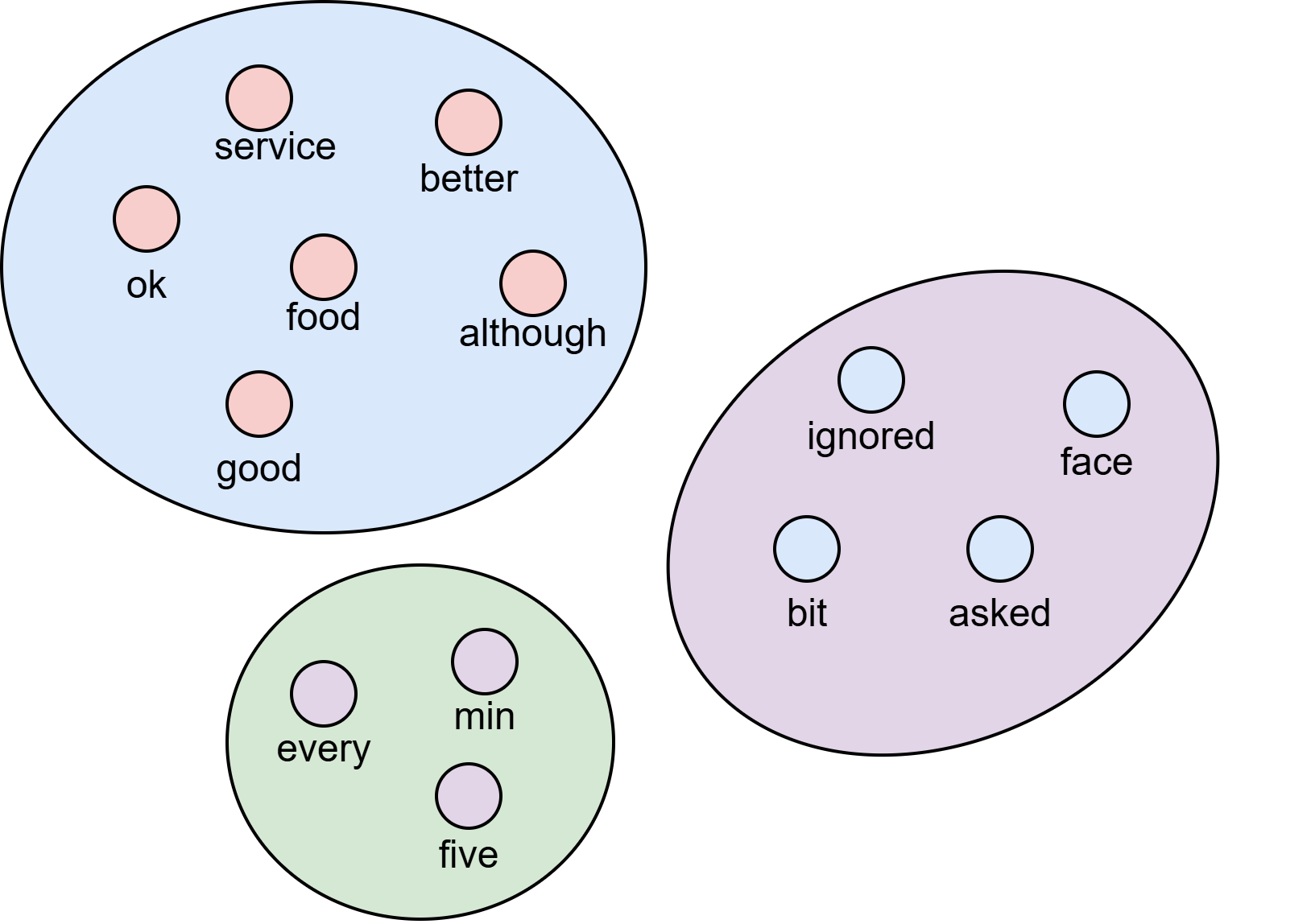}  
    \caption{Visualization of the hypergraph structure for the representative input, illustrating the model’s ability to form semantically coherent token groupings.}
    \label{fig:3}
\end{figure}

To explore the interpretability of the induced hypergraph structure, we analyze the model's output on a representative example: “\textit{service is good although a bit in your face, we were asked every five mins if food was ok, but better that than being ignored.}” Despite the absence of explicit syntactic supervision during hypergraph construction, the resulting hyperedges exhibit semantically meaningful groupings (Fig \ref{fig:3}). 

One hyperedge connects tokens such as \textit{bit, face, asked} and \textit{ignored}, all of which express negative sentiment or discomfort, capturing the notion of intrusive service. Another hyperedge clusters \textit{service, good, food, ok} and \textit{better}, reflecting a coherent positive sentiment span related to service and food quality. A third connects \textit{every, five, mins} aligning with a temporal expression. These patterns suggest that the model implicitly learns to organize tokens into functionally and semantically coherent structures, highlighting the capacity of the hypergraph to capture higher-order relational cues even without reliance on syntactic parsers. To further illustrate how these groupings emerge, we visualize the underlying token-wise clustering process in the form of a dendrogram (Fig.~\ref{fig:dendrogram-threshold}), where the cutoff threshold used to extract hyperedges is also marked. The dendrogram clearly reveals the relative proximity between tokens, highlighting where sharp transitions in similarity occur, informing how hyperedges are selectively formed.

To complement the qualitative analysis in Fig \ref{fig:3}, Table \ref{tab:hyperabsa-examples} presents additional examples of aspect-specific hyperedges extracted by HyperABSA across diverse sentences. Each row shows a sentence, its annotated aspect term, and the corresponding hyperedges learned by the model. These hyperedges group tokens not simply by syntactic adjacency, but by their functional or semantic role in relation to the aspect. For instance, in Example 1, the aspect \textit{cord} forms a hyperedge with tokens referring to charging, carrying, and battery life, reflecting how the model clusters tokens contributing to the user’s reasoning about portability. In Example 3, the aspect \textit{food} is linked to tokens describing quality, evaluation, and comparison, demonstrating the model’s ability to consolidate sentiment-bearing descriptors around the aspect. 

Across the examples, we also observe variation in the number and size of hyperedges extracted for each sentence. This is expected: because HyperABSA constructs hyperedges based on token-level similarity rather than a fixed syntactic template, the model dynamically allocates more or fewer hyperedges depending on how many semantically coherent token groups are present in the sentence. Interestingly, in nearly all examples, we see a consistent hyperedge containing punctuation marks. This reflects the model’s recognition that punctuation functions as structural boundary markers rather than contributing to aspect semantics, leading them to cluster together as a low-semantic, non-content group. We also see that [CLS] and [SEP] embeddings are frequently included within aspect-related hyperedges. Rather than being isolated, these special tokens act as global context anchors in the representation space, allowing the model to stabilize the similarity structure around the aspect term. Their presence suggests that the model is leveraging sentence-level contextualization to form hyperedges, rather than relying solely on local adjacency. Taken together, these patterns indicate that the hypergraph does not merely reflect surface structure, but instead reveals how the model organizes contextual, semantic, and functional relationships among tokens, including the learned roles of punctuation and transformer special tokens.

\begin{table}[htbp]
    \small
    \centering
    \setlength{\tabcolsep}{1mm}  
    \begin{tabular}{lcc|cc}
        \toprule
        \textbf{Model} & \multicolumn{2}{c|}{\textbf{Lap14}} & \multicolumn{2}{c}{\textbf{Rest14}} \\
        \cmidrule(lr){2-3} \cmidrule(lr){4-5}
                        & \textbf{Acc(\%)} & \textbf{F1(\%)} & \textbf{Acc(\%)} & \textbf{F1(\%)} \\
        \midrule
        ChatGPT (zero-shot)$^\ddagger$ & 77.64 & 72.30 & 82.39 & 73.64 \\
        ChatGPT (few-shot)$^\ddagger$ & 78.15 & 75.79 & 84.62 & 76.08 \\
        LLaMa-7B* & 75.37 & 71.63 & 80.04 & 70.97 \\
        LLaMa-13B* & 75.82 & 70.89 & 81.78 & 72.89 \\
        Alpaca-7B* & 74.58 & 71.18 & 80.38 & 71.36 \\
        Alpaca-13B* & 77.03 & 72.48 & 82.93 & 75.81 \\
        \textbf{HyperABSA} & \textbf{83.44} & \textbf{80.60} & \textbf{87.76} & \textbf{82.76} \\
        \bottomrule
    \end{tabular}
    \caption{Accuracy and F1 scores (\%) of HyperABSA compared with LLM variants on Lap14 and Rest14 datasets. The results with $^\ddagger$, * are retrieved from \cite{chen2024dynamic} and \cite{wang2024multivariate} respectively.} 
    \label{tab:llm-compare}
\end{table}

\begin{table}[t]
\small
\centering
\begin{tabular}{@{} c p{7cm} p{2.8cm} p{5.2cm} @{}}
\toprule
\textbf{No.} & \textbf{Sentence} & \textbf{Aspect} & \textbf{Hyperedges from HyperABSA} \\
\midrule

1 & I charge it at night and skip taking the cord with me because of the good battery life
& cord
& \begin{tabular}[t]{@{}p{5.2cm}@{}}
\raggedright\ttfamily
\texttt{['.', '[SEP]'],} \\
\texttt{['i', 'charge', 'at', 'night', 'and', 'skip', 'taking', 'me', 'because'],} \\
\texttt{['[CLS]', 'cord', 'with'],} \\
\texttt{['it', 'the', 'cord', 'of', 'the', 'good', 'battery', 'life']}
\end{tabular} \\ \\

2 & The tech guy then said the service center does not do 1-to-1 exchange and I have to direct my concern to the "sales" team, which is the retail shop which I bought my netbook from. 
& service center 
& \begin{tabular}[t]{@{}p{5.2cm}@{}}
\raggedright\ttfamily
\texttt{['[CLS]', 'tech', 'sales', 'team', 'retail', 'shop', 'bought', 'net', '\#\#book', 'from'],} \\
\texttt{['service', 'center],} \\
\texttt{['1', 'to', '1'],} \\
\texttt{['the', 'guy', 'then', 'said', 'and', '.'],} \\
\texttt{['does', 'not', 'do', 'exchange', 'have', 'to', 'direct', 'concern', 'to'],} \\
\texttt{['the', 'the', "'", "'", 'which', 'is', 'the', 'which'],} 
\end{tabular} \\ \\

3 & To be completely fair, the only redeeming factor was the food, which was above average, but couldn't make up for all the other deficiencies of Teodora 
& food 
& \begin{tabular}[t]{@{}p{5.2cm}@{}}
\raggedright\ttfamily
\texttt{['to', 'be', 'completely', 'fair'],} \\
\texttt{['the', 'only', 'red', '\#\#eem', '\#\#ing', 'factor', 'the'],} \\
\texttt{['was', 'food', 'which', 'was', 'above', 'average', 'but', 'couldn', 't', 'def', '\#\#iciencies', 'of', '\#\#ora', 'food'],} \\
\texttt{['the', 'only', 'red', '\#\#eem', '\#\#ing', 'factor', 'the'],} \\
\texttt{['for', 'all', 'the', 'other'],} \\
\texttt{['make', 'up'],} \\
\texttt{[',', ',', ',', "'", '.'],} \\
\texttt{['[CLS]', '[SEP]']} 
\end{tabular} \\ \\

4 & The food is uniformly exceptional, with a very capable kitchen which will proudly whip up whatever you feel like eating, whether it's on the menu or not 
& kitchen  

& \begin{tabular}[t]{@{}p{5.2cm}@{}}
\raggedright\ttfamily
\texttt{['the', 'is', 'uniformly', 'with', 'a', 'very', 'which', 'will'],} \\
\texttt{ ['food', 'exceptional', 'capable', 'kitchen', 'proudly', 'whip', 'up', 'eating', 'menu'],} \\
\texttt{['whatever', 'it', "'", 's', 'on', 'the'],} \\
\texttt{ ['you', 'feel', 'like'],} \\
\texttt{['whether', 'or', 'not'],} \\
\texttt{[',', ',', '.'],} \\
\texttt{ ['[CLS]', 'kitchen', '[SEP]'],} 
\end{tabular} \\ \\

5 & Staple entrees like moussaka more than make the grade, but the selection of clay pot-cooked dishes--tender lamb with orzo, or fish of the day with eggplant and zucchini--take comfort food to a new level. 
& entrees 

& \begin{tabular}[t]{@{}p{5.2cm}@{}}
\raggedright\ttfamily
\texttt{['\#\#s', 'the', 'selection', 'of', '-', '.', '\#\#s'],
 ['[CLS]', 'staple'],} \\
\texttt{ ['clay', 'pot', 'cooked', 'dishes', 'tender', 'lamb', 'fish', 'food'],} \\
\texttt{['like', 'grade', 'but', 'with', ',', 'or', 'with', 'and', 'comfort'],} \\
\texttt{ ['egg', '\#\#pl', '\#\#ant', 'zu', '\#\#cchi', '\#\#ni'],} \\
\texttt{['en', '\#\#tree', 'en', '\#\#tree','mo', '\#\#uss', '\#\#aka', 'or', '\#\#zo'],} \\
\texttt{['make', 'take', 'to', 'a', 'new', 'level'],}  \\
\texttt{  [',', '[SEP]',  '-', '-', '-', '-'],} 
\end{tabular} \\ \\

  

\bottomrule
\end{tabular}
\caption{Example sentences with aspect-related hyperedges extracted by HyperABSA.}
\label{tab:hyperabsa-examples}
\end{table}

\section*{Appendix D \quad Comparison with LLM Baselines}
\label{sec:appendixD}

Table~\ref{tab:llm-compare} reports a comparison between HyperABSA and several large language model (LLM) baselines on the Lap14 and Rest14 datasets. To ensure experimental fairness, all models operate on the same task formulation, dataset splits, and input text preprocessing as used in HyperABSA. 

On average across both datasets, HyperABSA consistently surpasses all LLM baselines. 
Compared to the strongest proprietary model - ChatGPT (few-shot) - HyperABSA achieves a mean improvement of +2.22\% in Accuracy and +2.60\% in F1. 
Relative to the best-performing open-source model, Alpaca-13B, the gains are even more substantial, with +3.63\% higher Accuracy and +4.89\% higher F1 on average. 

These improvements highlight that explicitly modeling token-level interactions as hyperedges enables more fine-grained sentiment reasoning than prompting-based inference in large autoregressive LMs. 
While LLMs excel at general language understanding, their implicit relational reasoning remains limited when aspect-level dependencies are subtle or cross-sentential. 
HyperABSA bridges this gap by learning structured relational inductive biases over token clusters, yielding superior precision in aspect-specific sentiment detection.

\section*{Appendix E \quad Sensitivity Analysis of \( \lambda\) and \( \rho \)}
\label{sec:appendixE}

\begin{figure}[htbp]
    \centering
    \includegraphics[width=0.7\columnwidth]{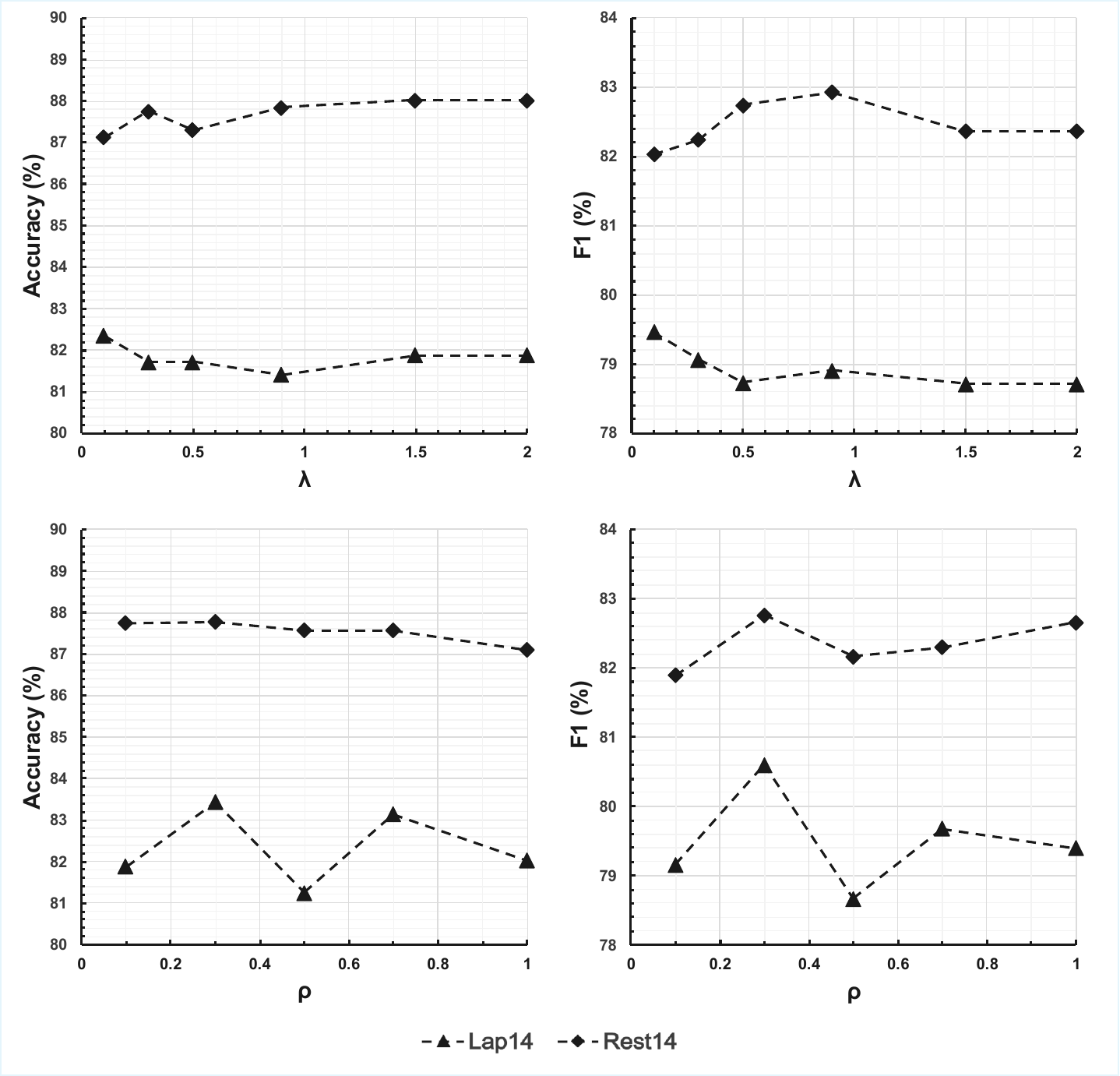}  
    \caption{Sensitivity analysis of the adaptive cutoff parameters $\lambda$ and $\rho$. The top row shows how Accuracy and F1 (\%) vary with $\lambda$, and the bottom row illustrates the effect of $\rho$.}
    \label{fig:lambda-rho}
\end{figure}

To evaluate the robustness of the adaptive cutoff mechanism in our hypergraph construction, we conduct a sensitivity analysis over the parameters $\lambda$ and $\rho$. The parameter $\lambda$ appears in Equation ~\ref{eq:fallback_criterion} where it scales the standard deviation of recent merge dissimilarities. The observed trend in the top row of Figure~\ref{fig:lambda-rho} reveals that as \(\lambda\) increases from 0 to approximately 0.5-1.0, both Accuracy and F1 score show a mild but consistent improvement before plateauing. This behavior reflects how \(\lambda\) governs the fallback cutoff sensitivity during hierarchical clustering: lower \(\lambda\) values make the model overly aggressive in merging clusters, which can conflate distinct aspect or opinion tokens into a single hyperedge. Such under-segmentation causes loss of fine-grained sentiment distinctions, slightly reducing classification performance. Conversely, when \(\lambda\) becomes too large, the cutoff becomes overly conservative, producing many small clusters that fragment coherent semantic groups, an instance of over-segmentation. The plateau observed beyond \(\lambda \approx 1.0\) suggests that the model’s adaptive cutoff mechanism effectively self-regulates, where further increases in \(\lambda\) have diminishing influence because the elbow-based criterion dominates in most cases. From this behavior, we can infer that a moderate \(\lambda\) allows the model to achieve a structurally balanced hypergraph, capturing both local token coherence and broader semantic grouping.

The parameter $\rho \in (0, 1]$ controls the fraction of recent merges considered when computing the second-order acceleration signal used to detect the elbow point. It effectively governs the temporal window of the clustering process, determining how local or global the merge compactness trends are when deciding where to cut the dendrogram. The bottom row of Figure~\ref{fig:lambda-rho} illustrates how the model’s performance varies with respect to the parameter \(\rho\). The stable performance observed for \(\rho\) between 0.2 and 0.4 indicates that the adaptive cutoff achieves its best balance when it integrates both local and global clustering information. At lower \(\rho\) values, the window of recent dissimilarities is too narrow, causing the model to rely heavily on fine-grained fluctuations in merge distances. This results in a noisy cutoff that can overemphasize short-range dependencies and generate excessively fragmented hyperedges. In contrast, higher \(\rho\) values expand the merge window excessively, smoothing out meaningful local variations and leading to overly coarse clusters that blur sentiment distinctions between aspects.

\end{document}